\documentclass{article}
\usepackage[utf8]{inputenc}

\usepackage{amsmath}
\usepackage{graphicx}

\newtheorem{theorem}{Theorem}
\newtheorem{definition}[theorem]{Definition}
\newtheorem{ax}[theorem]{Property}

\newcommand{\Bem}[1]{}
\newcommand{\figaddr}[1]{figs/#1}

\title{A Clustering Preserving Transformation for k-Means Algorithm Output}
\author{Mieczys{\l}aw A. K{\l}opotek }
\date{July 2022}

\newcommand{\wc}{0.3}

\begin{document}

\maketitle

\begin{abstract}
    This note introduces a novel clustering preserving transformation of cluster sets obtained from $k$-means algorithm. 
    This transformation may be used to generate new labeled data{}sets from existent ones. 
    It is more flexible that Kleinberg axiom based consistency transformation because data points in a cluster can be moved away and datapoints between clusters may come closer together. 
\end{abstract}

\section{Introduction}

In this note we introduce a novel clustering preserving transformation of cluster sets obtained from $k$-means algorithm.
It may be considered as a contribution towards formulation of clustering axiomatic system. 

From the practical point of view, this clustering preserving transformation can be used for purposes of:
\begin{itemize}
    \item generating new labeled datasets from existent ones, which may be of use in testing algorithms from $k$-means family in their stability on cluster perturbations which d not change the theoretical clustering,
    \item generating new labeled datasets from existent ones, obfuscating sensitive data  
\end{itemize}
From the theoretical standpoint, the contribution of this paper consists in proposing a less rigid cluster preserving transformation than centric consistency, known so far as the only cluster preserving transformation for $k$-means family of algorithms.  

As we will refer to the $k$-means algorithm, let us recall that it is aimed at minimizing  the cost function $Q$ (reflecting its quality in that the lower $Q$ the higher the quality)  of the form: 

\begin{equation} \label{eq:Q::kmeans}
Q(\Gamma)=\sum_{i=1}^m\sum_{j=1}^k u_{ij}\|\textbf{x}_i - \boldsymbol{\mu}_j\|^2
=\sum_{j=1}^k \frac{1}{n_j} \sum_{\{\mathbf{x}_i, \mathbf{x}_l\} \subseteq  C_j} \|\mathbf{x}_i - \mathbf{x}_l\|^2 
\end{equation} 
$$=\sum_{j=1}^k \frac{1}{2n_j} \sum_{\mathbf{x}_i \in C_j} 
\sum_{\mathbf{x}_l \in C_j} \|\mathbf{x}_i - \mathbf{x}_l\|^2 $$
for a data{}set $\mathbf{X}$
under some partition $\Gamma$ of the dataset $\mathbf{X}$ into the predefined number $k$ of  clusters (a cluster being a non-empty set, with empty intersections with other clusters), 
where  $u_{ij}$ is an indicator of the membership of data point $\textbf{x}_i$ in the cluster $C_j$ having the cluster center at $\boldsymbol{\mu}_j$ (which is the cluster's gravity center). 
Note that \cite{Pollard:1981} modified this definition by dividing the right hand side by $m$ in order to make comparable values for samples and the population, but we will only handle samples of a fixed size so this definition is sufficient for our purposes. 
A $k$-means algorithm finding exactly the clustering optimizing $Q$ shall be refereed to as $k$-means-ideal. 
There exist a number of  implementations of algorithms aiming at approximate optimization of the  $k$-means quality criterion.  For various versions of $k$-means algorithm see e.g. \cite{STWMAK:2018:clustering}.
Realistic implementations start from a randomized initial partition and then improve the $Q$ iteratively.  

Kleinberg \cite{Kleinberg:2002} introduced an axiomatic framework for distance-based clustering functions to which the widely used $k$-means algorithm belongs.

Two of the Kleinberg's axioms represent transformations under which the clustering should be preserved. This can be viewed as a way to create new labelled data sets for testing of clustering functions. 
Regrettably, the $k$-means algorithm (one of the most frequently used clustering algorithms) fails to match requirements of this axiomatic system, it fails on the consistency axiom. 

The consistency axiom of Kleinberg has the following form:

\begin{ax}  \label{ax:consistency} \cite{Kleinberg:2002}
Let $\Gamma$ be a partition of $S$, and $d$ and
$d'$  two distance functions on $S$. We say that $d'$ 
 is a $\Gamma$-transformation of $d$ ($\Gamma(d)=d')$ if (a) for
all $i,  j \in  S$ belonging to the same cluster of 
$\Gamma$, we have $d'(i, j) \le d(i, j)$   and (b) for
all $i,  j \in  S$ belonging to different clusters of $\Gamma$,
 we have $d'(i, j) \ge d(i, j)$.
  The clustering function 
  $f$ has the \emph{consistency} property 
if for each distance function $d$ and its $\Gamma$-transform   $d'$ the following holds: if $f(d) =\Gamma$,  then $f(d') = \Gamma$%
\end{ax}

The paper \cite{Klopotek:2022continuous} suggested a substitute for the consistency axiom, proposing so-called centric consistency.

\begin{definition} \cite{Klopotek:2022continuous}
Let $\mathcal{E}$ be an embedding of the data{}set $S$ with distance function $d$ (induced by this embedding). 
Let $\Gamma$ be a partition of this data{}set.
Let $C\in \Gamma$ and let $\boldsymbol\mu_c$ be the gravity center of the cluster $C$ in $\mathcal{E}$.
We say that we execute the \emph{$\Gamma^*$} transformation   $\Gamma^*(d;\lambda)=d'$) if for some $0<\lambda\le 1$ a new embedding $\mathcal{E}'$ is created in which  each element $x$ of $C$, with coordinates $\mathbf{x}$ in $\mathcal{E}$, has coordinates $\mathbf{x'}$ in $\mathcal{E}$  such that   $\textbf{x'}=\boldsymbol\mu_c+\lambda(\textbf{x}-\boldsymbol\mu_c)$, all coordinates of all other data points are identical, and $d'$ is induced by $\mathcal{E}'$.  $C$ is then said to be subject of the centric transform.
\end{definition} 

Note that the set of possible  $\Gamma^*$-transformations for a given partition is neither a subset nor super{}set of the set of possible Klein{}berg's $\Gamma$-transformation in general.

\begin{ax}\label{ax:centricconsistency} \cite{Klopotek:2022continuous}
A clustering method has the property of \emph{centric consistency}
if after a $\Gamma^*$ transform it returns the same partition. 
\end{ax}

\begin{theorem}{\cite{MAKRAK:2020:ICAISC2020}} \label{thm:globalCentricCinsistencyFor2means}
$k$-means algorithm satisfies centric consistency property in the following way:  
if the partition $\Gamma$ of the set $S$ with distances $d$ is a global minimum of $k$-means, and $k=2$, and the partition $\Gamma$  has been subject to $\Gamma^*$-transformation  yielding distances $d'$,  then $\Gamma$ is also a global minimum of $k$-means under distances $d'$.  
\end{theorem}
See proof in paper \cite{MAKRAK:2020:ICAISC2020}. 
A more general theorem was introduced in 

\begin{theorem}{\cite[Theorem 34]{Klopotek:2022continuous}} \label{thm:globalCentricCinsistencyForKmeans} 
$k$-means algorithm    satisfies
 centric consistency in the following way:  
if the partition $\Gamma$ is a global minimum of $k$-means, 
and the partition $\Gamma$  has been subject to $\Gamma^*$-transform yielding $\Gamma'$, then $\Gamma'$ is also a global minimum of $k$-means.
\end{theorem}

For the proof of this theorem see \cite[Theorem 34]{Klopotek:2022continuous}.

The axiom/property \ref{ax:centricconsistency} may be considered as a rigid one because all data points of a cluster are transformed linearly. 

Therefore in this note we elaborate another clustering preserving transformation suitable for $k$-means algorithm so that it can be used with this category of algorithms. 

It is worth noting that this transformation differs from Kleinberg's $\Gamma$ transformation. which required that distances within a cluster are reduced and distances between clusters are increased. 
With our transformation it is valid to increase distances within a cluster and decrease distances between clusters. 
This fact may be considered a hint that there is a need  to redefine our formal cluster understanding. 
\begin{definition}
For a point set $P$, the \emph{centric set transform} by factor $\lambda$ 
is a transformation assigning each point $\mathbf{x} \in P$ a data{}point $\mathbf{x'}$ such that $\mathbf{x'}=\lambda(\mathbf{x}-\boldsymbol\mu(P))+ \boldsymbol\mu(P)$
where $\boldsymbol\mu(P)$ is the gravity center of the data{}set $P$. 
\end{definition}

\begin{figure}
\begin{center}
\includegraphics[width=0.8\textwidth]{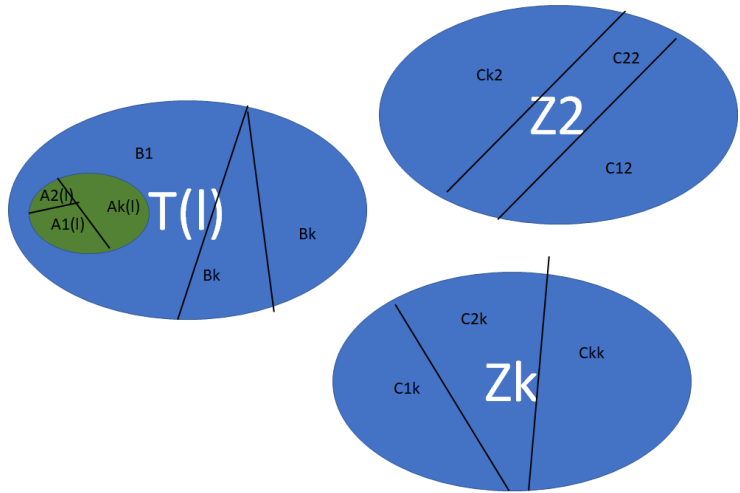} 
 \end{center}
\caption{Illustration of the task 
}\label{fig:theTask}
\end{figure}

In the next Section \ref{sec:related}, we provide a brief overview of various areas where cluster preserving transformations of data are an issue. 
Then we formulate our proposal of cluster preserving transformation for $k$-means in Section \ref{sec:ourcontribution} and prove analytically its correctness. 
In Section \ref{sec:examples} we illustrate the operation of the transformation with some examples. 
Section \ref{sec:final} contains some final remarks. 

\section{Related Work}\label{sec:related}

The most heavily cited (and at the same time contradictory) axiomatic framework for clustering was introduced by Klein{}berg 
\cite{Kleinberg:2002}. 
An overview of various efforts to overcome the problems of that framework is provided e.g. in \cite{Klopotek:2022continuous}. 

In this overview of related work let us concentrate therefore on the aspect of cluster-preserving transformations which are often related to diverse practical applications.  
Note that in \cite{Klopotek:2022continuous}   also the applicability to the problem of test{}bed creation for clustering algorithms from the $k$-means family is addressed.  

Roth et al. \cite{Roth:2003} investigated the issue of preservation of clustering when embedding non-euclidean data into the Euclidean space. 
They showed that clustering functions, that remain invariant under additive shifts of the pairwise proximities, can be reformulated as clustering problems in Euclidean spaces. 

Parameswaran and   Blough
\cite{Parameswaran:2005}
considered the issue of  cluster preserving transformations from the point of view of privacy preserving. They designed a 
Nearest Neighbor Data Substitution (NeNDS), a new  data obfuscation technique with  strong privacy-preserving properties while  maintaining data clusters.
Cluster preserving transformations with the property of 
privacy preserving focusing on the $k$-means algorithm are investigated by Ramírez and  Auñón \cite{Ramirez:2020privacy}.
Privacy preserving methods for various $k$-means variants boosted to large scale data are further elaborated in 
\cite{Gao:2017}. 
Keller et al. \cite{Keller:2021} investigate such transformations for other types of clustering algorithms. 
A thorough survey of  
privacy-preserving clustering for big data
can be found in \cite{Zhao:2020} by Zhao et al.

Howland  and  Park
\cite{Howland:2007} 
proposed  models incorporating prior knowledge about the existing structure and developed for them  dimension reduction methods independent of the original  term-document matrix dimension.
Other, more common dimensionality reduction methods for clustering (including PCA and Laplacian embedding) are reviewed  by 
Ding \cite{Ding:2009}. 

Larsen et al. 
\cite{Larsen:2016heavy}
reformulate the heavy hitter problem of stream mining in terms of a clustering problem and elaborate algorithms fulfilling the requirement of "cluster preserving clustering". 

Zhang et al. 
\cite{Zhang:2019}
developed clustering structure preserving transformations for graph streaming data, when there is a need to sample the graph.

\section{A Clustering Preserving Transformation}\label{sec:ourcontribution}

\begin{theorem}{} \label{thm:globalCentricConsistencyForKmeansSubsetted} 
Let a partition $\Gamma_o=\{T,Z_2,\dots,Z_k\}$ be an optimal partition of a dataset $\mathbf{X}$ under 
$k$-means algorithm, that is minimizing $Q(\Gamma)$ from eq. \ref{eq:Q::kmeans} over all $\Gamma$ with $card(\Gamma)=card(\Gamma_o)$.
Let a subset $P$ of $T$ be subjected to centric set transform 
yielding $P'(\lambda)$ 
and $T'(\lambda)=(T-P)\cup P'(\lambda)$. Then  
partition $\{T',Z_2,\dots,Z_k\}$ is an optimal partition of $T'\cup Z_2\cup \dots \cup Z_k$ under $k$-means. 
    \end{theorem}

\begin{definition}
In the above Theorem \ref{thm:globalCentricConsistencyForKmeansSubsetted},
$\{T',Z_2,\dots,Z_k\}$ shall be called
\emph{$\Gamma^{++}$ transform} of 
$\{T,Z_2,\dots,Z_k\}$ 
\end{definition}

\begin{figure}
\begin{center}
\includegraphics[width=0.8\textwidth]{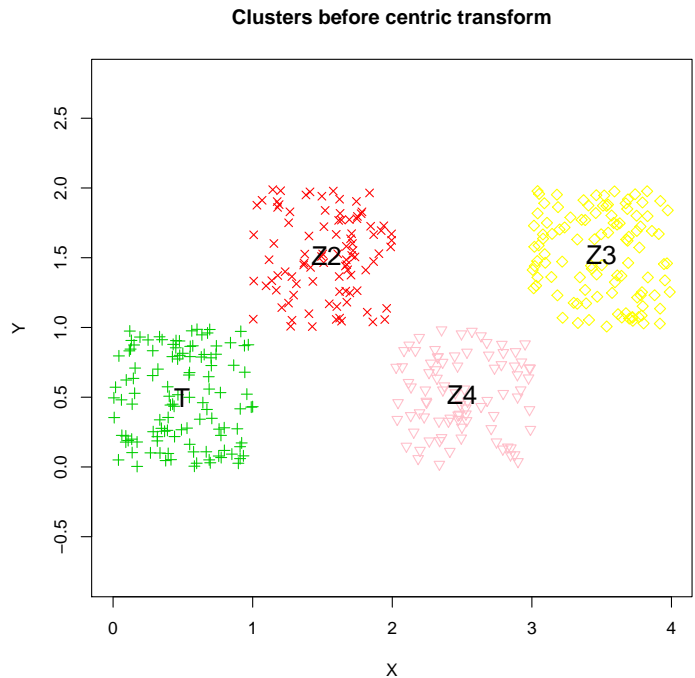} 
 \end{center}
\caption{Example clusters before centric transformation 
}\label{fig:clbeforect}
\end{figure}

  (Outline of the proof) 
Let the optimal clustering for a given set of objects $\mathbf{X}$ consist of $k$ clusters: $T$ and $Z_2,\dots,Z_k$. 
Let $T$ consist of two disjoint subsets $P$, $Y$,
$T=P \cup Y$ and let us ask the question whether or not centric transform of the set $P$ will affect the optimality of clustering. 
Let   $T'(\lambda)=P'(\lambda) \cup Y$ with $P'(\lambda)$ being an image of $P$ under centric set transformation. 
See Figure \ref{fig:theTask}

The cluster centre of $T'(\lambda)$ will be the same as that of $T$. 
 We ask if $\{T'(\lambda),Z_2,\dots,Z_k\}$ is the globally optimal clustering of $T'(\lambda)\cup Z_2\cup \dots \cup Z_k$.
Assume the contrary, that is that there exists 
a clustering into 
sets $K_i'(\lambda)=A_i'(\lambda)\cup B_i\cup C_{i,2}\cup \dots \cup C_{i,k}$, $i=1,\dots,k$  where $P=A_1\cup A_2\dots \cup A_k$, and 
$A'_i(\lambda)$ are the points obtained from $A_i$ when $P$ is subjected to centric set transformation. hence 
$P'(\lambda)=A'_1(\lambda)\cup \dots \cup A'_k(\lambda), Y=B_1\cup\dots\cup B_k, Z_j=C_{1,j}\cup \dots \cup  C_{k,j}$,
 that, for some $\lambda=\lambda^*\in (0,1)$  has lower clustering quality 
function value $Q(\{K'_1(\lambda),\dots,K_k'(\lambda)\} )$.
Define also the function 
$h(\lambda)=Q(\{T'(\lambda),Z_2,Z_k\})-Q(\{K_1'(\lambda),\dots,K_k'(\lambda)\})$.
Due to optimality assumption, $h(1)\le 0$.  
\Bem{
1/2 |C| sum_x \in C sum_Y \in C  (x-y)^2
= in case of two 
1/2 (n1+n2)( n1*n2(mu1-mu2)^2+ n2*n1(mu2-mu1)^2
= in case of three 
1/2 (n1+n2+n3)( n1*n2(mu1-mu2)^2+ n1*n3(mu1-mu3)^2+
n2*n1(mu2-mu1)^2
n2*n3(mu2-mu3)^2
n3*n1(mu3-mu1)^2
n3*n2(mu3-mu2)^2

)

} 

%

\begin{figure}
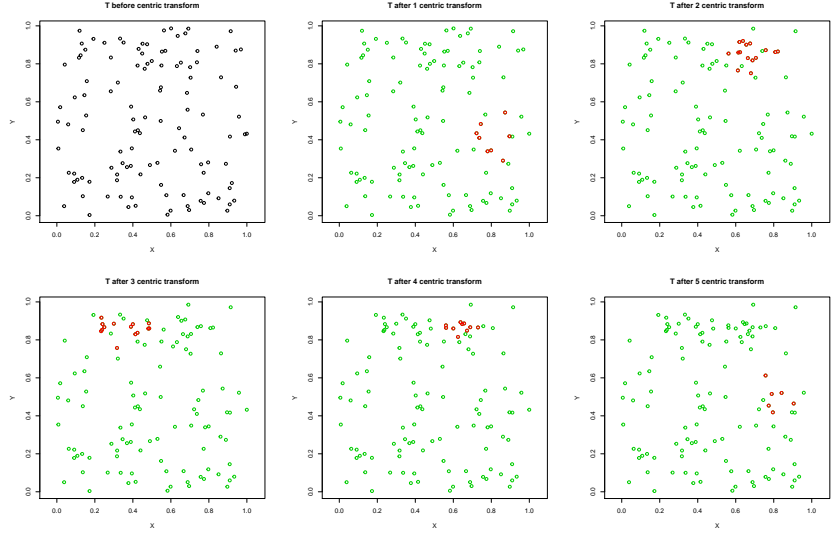

\begin{center}
\includegraphics[width=\wc\textwidth]{\figaddr{T_before_centric_transform.pdf}}
\includegraphics[width=\wc\textwidth]{\figaddr{T_after_1_centric_transform.pdf}} 
\includegraphics[width=\wc\textwidth]{\figaddr{T_after_2_centric_transform.pdf}} 
\includegraphics[width=\wc\textwidth]{\figaddr{T_after_3_centric_transform.pdf}} 
\includegraphics[width=\wc\textwidth]{\figaddr{T_after_4_centric_transform.pdf}} 
\includegraphics[width=\wc\textwidth]{\figaddr{T_after_5_centric_transform.pdf}} 
 \end{center}
\caption{Example: $T$ cluster before centric transformation, and then after 1,2,3,4 and 5 centric transformations. 
}\label{fig:Tbeforeafter12345ct}
\end{figure}

Let us discuss now the centric set transform with   $\lambda=0$ (being a notation abuse).
In this case all points from all $A'_i(0)$  collapse to a single point. This point can be closer to  one of  $\boldsymbol\mu(K'_{i*}(0))$  then from the other. Assume they are closer to $\boldsymbol\mu(K_{i*}'(0))$. 
In this case 
$Q(\{K_1"(\lambda),\dots,K_k"(\lambda)\}) \le
Q(\{K_1'(\lambda),\dots,K_k'(\lambda)\})$
for $\lambda=0$
where 
$K_{i*}"(\lambda)=P'(\lambda)\cup B_{i*}\cup C_{i*,2} \cup \dots \cup C_{i*,k}$,  and
$K_{i}"(\lambda)=P'(\lambda)\cup B_{i}\cup C_{i,2} \cup \dots \cup C_{i,k}$, for $i$ different from $i*$. 
As all points subject to centric set transform are contained in a single set, we get
\begin{align*}
&Q(\{T'(0),Z_2,\dots,Z_k\})-Q(\{K_1"(0),\dots,K_k"(0)\})=
\\ &
=Q(\{T'(1),Z_2,\dots,Z_k\})-Q(\{K"(1),\dots,K_k"(1)\} )
\le 0    
\end{align*}
because $Q(\{T'(1),Z_2,\dots,Z_k\})=Q(\{T,Z_2,\dots,Z_k\})$ is the optimum. 
Hence also 
$$Q(\{T'(0),Z_2,\dots,Z_k\})-Q(\{K'(0),\dots,K_k'(0)\} )
\le 0$$ that is $h(0)\le 0$. 
  
It is also easily seen that $h(\lambda)$ is a quadratic function of $\lambda$. 
This can be seen as follows:
\begin{align*}
& Q(\{T'(\lambda),Z_2,\dots,Z_k\})=
\\=&
\left(\sum_{\mathbf{x}\in T'(\lambda)} \|\mathbf{x}-\boldsymbol\mu( T'(\lambda))\|^2\right)
+
\sum_{j=2}^k\left(\sum_{\mathbf{x}\in Z_j} \|\mathbf{x}-\boldsymbol\mu( Z_j) \|^2\right)
\\=&
\left(\sum_{\mathbf{x}\in P'(\lambda)} \|\mathbf{x}-\boldsymbol\mu( P'(\lambda))\|^2\right)
+
\left(\sum_{\mathbf{x}\in Y} \|\mathbf{x}-\boldsymbol\mu( Y)\|^2\right)
\\&+\|\boldsymbol\mu(P'(\lambda))-\boldsymbol\mu(Y)\|^2\cdot \frac{1}{1/|P|+1/|Y|}
+ 
\sum_{j=2}^k\left(\sum_{\mathbf{x}\in Z_j} \|\mathbf{x}-\boldsymbol\mu( Z_j) \|^2\right)
\\=&
\sum_{i=1}^k\left(\sum_{\mathbf{x}\in A'_i(\lambda)} \|\mathbf{x}-\boldsymbol\mu( A'_i(\lambda))\|^2\right)
+\sum_{i=1}^k|A_i|\|\boldsymbol\mu( A'_i(\lambda))-\boldsymbol\mu( P'(\lambda))\|^2
\\&+
\left(\sum_{\mathbf{x}\in Y} \|\mathbf{x}-\boldsymbol\mu( Y)\|^2\right)
+\|\boldsymbol\mu(P'(\lambda))-\boldsymbol\mu(Y)\|^2\cdot \frac{1}{1/|P|+1/|Y|}
\\&+
\left(\sum_{\mathbf{x}\in Z} \|\mathbf{x}-\boldsymbol\mu( Z) \|^2\right)
\end{align*}

This expression is obviously quadratic in $\lambda$, since each point $\mathbf{x}\in P$  is transformed linearly to $\boldsymbol\mu(P)+\lambda(\mathbf{x}-\boldsymbol\mu(P))$. 
Note that $\boldsymbol\mu( P'(\lambda))=\boldsymbol\mu( P  )$
so it does not depend on $\lambda$. 
On the other hand 
\begin{align*}
&Q(\{K_1'(\lambda),\dots,K_k'(\lambda)\})=
\\&=
Q(\{
A_1'(\lambda)\cup B_1\cup C_{1,2}\cup \dots \cup C_{1,k}
,\dots,
A_k'(\lambda)\cup B_k\cup C_{i,2}\cup \dots \cup C_{k,k}
\})
\\&=
\sum_{i=1}^k \left(\sum_{\mathbf{x}\in A'_i(\lambda)} \|\mathbf{x}-\boldsymbol\mu( A'_i(\lambda))\|^2\right)
+\sum_{i=1}^k\left(\sum_{\mathbf{x}\in B_i} \|\mathbf{x}-\boldsymbol\mu( B_i)\|^2\right)
\\&+\sum_{i=1}^k\sum_{j=2}^k\left(\sum_{\mathbf{x}\in C_{i,j}} \|\mathbf{x}-\boldsymbol\mu( C_{i,j})\|^2\right)
\\&+\sum_{i=1}^k\frac{1}{|A_i|+|B_i|+\sum_{j=2}^k|C_{i,j}|}
\biggl(
|A_i||B_i| \|\boldsymbol\mu(A'_i(\lambda))-\boldsymbol\mu(B_i)\|^2
\\&
+|A_i|\sum_{j=2}^k|C_{i,j}| \|\boldsymbol\mu(A'_i(\lambda))-\boldsymbol\mu(C_{i,j})\|^2
+|B_i|\sum_{j=2}^k|C_{i,j}| \|\boldsymbol\mu(B_i)-\boldsymbol\mu(C_{i,j})\|^2
\\&
+\sum_{j=2}^{k-1}|C_{i,j}|\sum_{j'=j+1}^{k}|C_{i,j'}| \|\boldsymbol\mu(C_{i,j})-\boldsymbol\mu(C_{i,j'})\|^2
\biggr)
\end{align*}

Then 

\begin{align*}
&h(\lambda)=
Q(\{T'(\lambda),Z_2,\dots,Z_k\})
- Q(\{K_1'(\lambda),\dots,K_k'(\lambda)\})
\\=&
\sum_{i=1}^k\left(\sum_{\mathbf{x}\in A'_i(\lambda)} \|\mathbf{x}-\boldsymbol\mu( A'_i(\lambda))\|^2\right)
+\sum_{i=1}^k|A_i|\|\boldsymbol\mu( A'_i(\lambda))-\boldsymbol\mu( P'(\lambda))\|^2
\\&+
\left(\sum_{\mathbf{x}\in Y} \|\mathbf{x}-\boldsymbol\mu( Y)\|^2\right)
\\&+\|\boldsymbol\mu(P'(\lambda))-\boldsymbol\mu(Y)\|^2\cdot \frac{1}{1/|P|+1/|Y|}
+
\left(\sum_{\mathbf{x}\in Z} \|\mathbf{x}-\boldsymbol\mu( Z) \|^2\right)
\\&
-\sum_{i=1}^k \left(\sum_{\mathbf{x}\in A'_i(\lambda)} \|\mathbf{x}-\boldsymbol\mu( A'_i(\lambda))\|^2\right)
-\sum_{i=1}^k\left(\sum_{\mathbf{x}\in B_i} \|\mathbf{x}-\boldsymbol\mu( B_i)\|^2\right)
\\&-\sum_{i=1}^k\sum_{j=2}^k\left(\sum_{\mathbf{x}\in C_{i,j}} \|\mathbf{x}-\boldsymbol\mu( C_{i,j})\|^2\right)
\\&-\sum_{i=1}^k\frac{1}{|A_i|+|B_i|+\sum_{j=2}^k|C_{i,j}|}
\biggl(
|A_i||B_i| \|\boldsymbol\mu(A'_i(\lambda))-\boldsymbol\mu(B_i)\|^2
\\&
+|A_i|\sum_{j=2}^k|C_{i,j}| \|\boldsymbol\mu(A'_i(\lambda))-\boldsymbol\mu(C_{i,j})\|^2
+|B_i|\sum_{j=2}^k|C_{i,j}| \|\boldsymbol\mu(B_i)-\boldsymbol\mu(C_{i,j})\|^2
\\&
+\sum_{j=2}^{k-1}|C_{i,j}|\sum_{j'=j+1}^{k}|C_{i,j'}| \|\boldsymbol\mu(C_{i,j})-\boldsymbol\mu(C_{i,j'})\|^2
\biggr)
\end{align*}
\begin{align*}
\\=&
\sum_{i=1}^k|A_i|\|\boldsymbol\mu( A'_i(\lambda))-\boldsymbol\mu( P'(\lambda))\|^2
\\&-\sum_{i=1}^k\frac{1}{|A_i|+|B_i|+\sum_{j=2}^k|C_{i,j}|}
\biggl(
|A_i||B_i| \|\boldsymbol\mu(A'_i(\lambda))-\boldsymbol\mu(B_i)\|^2
\\&
+|A_i|\sum_{j=2}^k|C_{i,j}| \|\boldsymbol\mu(A'_i(\lambda))-\boldsymbol\mu(C_{i,j})\|^2
\biggr)
\\&+
\left(\sum_{\mathbf{x}\in Y} \|\mathbf{x}-\boldsymbol\mu( Y)\|^2\right)
\\&+\|\boldsymbol\mu(P'(\lambda))-\boldsymbol\mu(Y)\|^2\cdot \frac{1}{1/|P|+1/|Y|}
+
\left(\sum_{\mathbf{x}\in Z} \|\mathbf{x}-\boldsymbol\mu( Z) \|^2\right)
\\&
-\sum_{i=1}^k\left(\sum_{\mathbf{x}\in B_i} \|\mathbf{x}-\boldsymbol\mu( B_i)\|^2\right)
-\sum_{i=1}^k\sum_{j=2}^k\left(\sum_{\mathbf{x}\in C_{i,j}} \|\mathbf{x}-\boldsymbol\mu( C_{i,j})\|^2\right)
\\&-\sum_{i=1}^k\frac{1}{|A_i|+|B_i|+\sum_{j=2}^k|C_{i,j}|}
\biggl(
|B_i|\sum_{j=2}^k|C_{i,j}| \|\boldsymbol\mu(B_i)-\boldsymbol\mu(C_{i,j})\|^2
\\&
+\sum_{j=2}^{k-1}|C_{i,j}|\sum_{j'=j+1}^{k}|C_{i,j'}| \|\boldsymbol\mu(C_{i,j})-\boldsymbol\mu(C_{i,j'})\|^2
\biggr)
%
\end{align*}
and keeping in mind that $\boldsymbol\mu( P'(\lambda))=\boldsymbol\mu( P )$
\begin{align*}
\\=&
\sum_{i=1}^k|A_i|\|\boldsymbol\mu( A'_i(\lambda))-\boldsymbol\mu( P)\|^2
\\&-\sum_{i=1}^k\frac{1}{|A_i|+|B_i|+\sum_{j=2}^k|C_{i,j}|}
\biggl(
|A_i||B_i| \|\boldsymbol\mu(A'_i(\lambda))-\boldsymbol\mu(B_i)\|^2
\\&
+|A_i|\sum_{j=2}^k|C_{i,j}| \|\boldsymbol\mu(A'_i(\lambda))-\boldsymbol\mu(C_{i,j})\|^2
\biggr)
%
\\&+c_h
\end{align*}


\noindent 
 where $c_h$  is a constant (independent of $\lambda$ and all the $\boldsymbol\mu$s that depend on $\lambda$, are linearly dependent on it (by definition of centric set transform). 
 \begin{align*}
\\c_h=&
\left(\sum_{\mathbf{x}\in Y} \|\mathbf{x}-\boldsymbol\mu( Y)\|^2\right)
\\&+\|\boldsymbol\mu(P)-\boldsymbol\mu(Y)\|^2\cdot \frac{1}{1/|P|+1/|Y|}
+
\left(\sum_{\mathbf{x}\in Z} \|\mathbf{x}-\boldsymbol\mu( Z) \|^2\right)
\\&
-\sum_{i=1}^k\left(\sum_{\mathbf{x}\in B_i} \|\mathbf{x}-\boldsymbol\mu( B_i)\|^2\right)
-\sum_{i=1}^k\sum_{j=2}^k\left(\sum_{\mathbf{x}\in C_{i,j}} \|\mathbf{x}-\boldsymbol\mu( C_{i,j})\|^2\right)
\\&-\sum_{i=1}^k\frac{1}{|A_i|+|B_i|+\sum_{j=2}^k|C_{i,j}|}
\biggl(
|B_i|\sum_{j=2}^k|C_{i,j}| \|\boldsymbol\mu(B_i)-\boldsymbol\mu(C_{i,j})\|^2
\\&
+\sum_{j=2}^{k-1}|C_{i,j}|\sum_{j'=j+1}^{k}|C_{i,j'}| \|\boldsymbol\mu(C_{i,j})-\boldsymbol\mu(C_{i,j'})\|^2
\biggr)
%
\end{align*}

\begin{figure}
\begin{center}%
\includegraphics[width=0.6\textwidth]{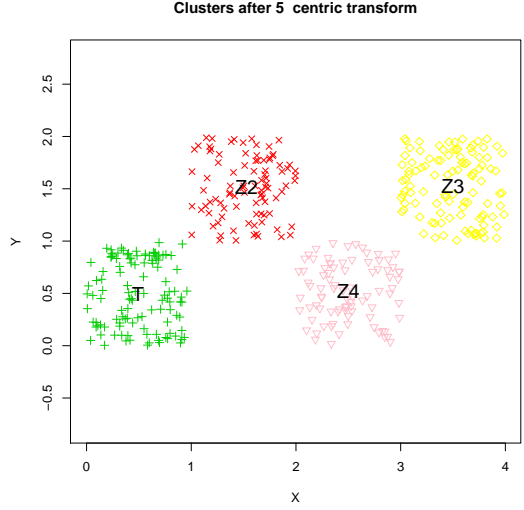} 
 \end{center}%
\caption{Example clusters after 5 centric transformations
of data from Fig. \ref{fig:clbeforect}
}\label{fig:clafter5ct}
\end{figure}
 
\begin{figure}
\begin{center}
\includegraphics[width=0.6\textwidth]{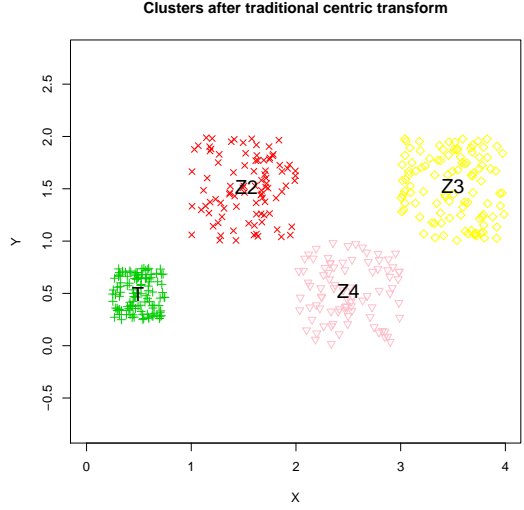} 
 \end{center}
\caption{Example clusters after traditional $\Gamma^*$ transformation 
of data from Fig. \ref{fig:clbeforect}
}\label{fig:claftertradct}
\end{figure}

 Recall that  
 $\boldsymbol\mu( A'_i(\lambda))-\boldsymbol\mu( P )
 =1/|A_i|\sum_{\mathbf{x}\in A}
 \lambda(\mathbf{x}-\boldsymbol\mu( P )) 
 =
 \lambda \mathbf{v_{A_i}}
 $,
 where $ \mathbf{v_{A_i}}=\boldsymbol\mu( A_i)-\boldsymbol\mu( P )$ is a vector independent of $\lambda$. 
 Hence 
 $\|\boldsymbol\mu( A'_i(\lambda))-\boldsymbol\mu( P )\|^2=\lambda^2  \mathbf{v_{A_i}}^T \mathbf{v_{A_i}}
 $. 
 Similarly 
 \begin{align*}
 \|\boldsymbol\mu&(A'_i(\lambda))-\boldsymbol\mu(B_i)\|^2
= 
 \|(\boldsymbol\mu(A'_i(\lambda)
 -\boldsymbol\mu( P ))
 +(\boldsymbol\mu( P )
 -\boldsymbol\mu(B_i))\|^2
 \\ & = 
 \|\boldsymbol\mu(A'_i(\lambda)
 -\boldsymbol\mu( P )\|^2
 +\|\boldsymbol\mu( P )
 -\boldsymbol\mu(B_i)\|^2+
 2(\boldsymbol\mu(A'_i(\lambda)
 -\boldsymbol\mu( P ))^T(\boldsymbol\mu( P )
 -\boldsymbol\mu(B_i))
 \\ & = 
 \lambda^2  \mathbf{v_{A_i}}^T \mathbf{v_{A_i}}
 +\|\boldsymbol\mu( P )
 -\boldsymbol\mu(B_i)\|^2+
 2 \lambda \mathbf{v_{A_i}}^T(\boldsymbol\mu( P )
 -\boldsymbol\mu(B_i))
 \\ & = \lambda^2  \mathbf{v_{A_i}}^T \mathbf{v_{A_i}}
 + \lambda c_{A_iB_iP}
 +c_{B_iP}
 \end{align*}
 with $ c_{A_iB_iP}= 2 \mathbf{v_{A_i}}^T(\boldsymbol\mu( P )
 -\boldsymbol\mu(B_i)),
 c_{B_iP}=\|\boldsymbol\mu( P )
 -\boldsymbol\mu(B_i)\|^2$ being constants independent of $\lambda$, 
 whereby only the first summand depends on $\lambda^2$. 
 Similarly 
$$ \|\boldsymbol\mu(A'_i(\lambda))-\boldsymbol\mu(C_{i,j})\|^2
= \lambda^2  \mathbf{v_{A_i}}^T \mathbf{v_{A_i}}
 + \lambda c_{A_iC_{i,j}P}
 +c_{C_{i,j}P}
 $$
 with $ c_{A_iC_{i,j}P},
 c_{C_{i,j}P}$ being constants independent of $\lambda$, 
 and so on. 
 Therefore we
 can rewrite 
 the 
 $h(\lambda)$ as 
 \begin{align*}
  h(\lambda)&=   
  \sum_{i=1}^k 
  |A_i| \lambda^2  \mathbf{v_{A_i}}^T \mathbf{v_{A_i}}
  \\&-\sum_{i=1}^k\frac{1}{|A_i|+|B_i|+\sum_{j=2}^k|C_{i,j}|}
\biggl(
|A_i||B_i| 
(\lambda^2  \mathbf{v_{A_i}}^T \mathbf{v_{A_i}}
 + \lambda c_{A_iB_iP}
 +c_{B_iP})
\\&
+|A_i|\sum_{j=2}^k|C_{i,j}| 
(\lambda^2  \mathbf{v_{A_i}}^T \mathbf{v_{A_i}}
 + \lambda c_{A_iC_{i,j}P}
 +c_{C_{i,j}P}
 )
\biggr)
+c_h
 \end{align*}

 So the coefficient at $\lambda^2$ amounts to: 
 $$
 \sum_{i=1}^k
 \left(|A_i|-
 \frac{|A_i|(|B_i|+\sum_{j=2}^k|C_{i,j}|)}{|A_i|+|B_i|+\sum_{j=2}^k|C_{i,j}|}
 \right)  \mathbf{v_{A_i}}^T \mathbf{v_{A_i}}
 $$
 which is bigger than 0 if only $|A_i|>0$.
 which is the case by our assumption of an alternative clustering. 
 Therefore, for $\lambda$ large enough, $h(\lambda)>0$.

   As $h(\lambda)$ is a quadratic function in $\lambda$, and   $h(0)\le 0$ and $h(1)\le 0$, then also $h(\lambda)\le 0$ for any value of $\lambda$ between 0 and 1.  
This completes the proof. 

\section{Examples}\label{sec:examples}

Consider the data and clusters from Figure \ref{fig:clbeforect}. 
The left top picture from Figure \ref{fig:Tbeforeafter12345ct} shows the $T$ cluster from Figure \ref{fig:clbeforect}. 
Then you see in other pictures from Figure \ref{fig:Tbeforeafter12345ct} results of changing $T$ by   a series of centric set transformations of some fragments of $T$. 
In spite of the fact, that the data was transformed, the overall shape of the cluster did not change much and no systematic changes can be observed. This means that the proposed transformation, which preserves the theoretical clustering via $k$-means-ideal, can be used as a way to verify the susceptibility of a real algorithm to slight changes in data positions.
In the end, in Figure \ref{fig:clafter5ct} the result of transforming data from \ref{fig:clbeforect} via 5 centric set transformations of fragments of $T$.

The previously proposed $\Gamma^*$ transformation, which also keeps the theoretical $k$-means clustering, transforms the data more rigidly, as you see in figure \ref{fig:claftertradct} - the cluster $T$ increases its gap separating it from other clusters under this transformation.

A next experiment was to compare 
  $\Gamma$ with $\Gamma^{++}$. An artificial dataset consisting of 10000 datapoints in 3D uniformly distributed over two  squares of same edge, touching each other at a corner point. They were transformed increasing angular distance to the diagonal by factor 1.9. Then either a    $\Gamma$ or  $\Gamma^{++}$ was applied to one of the 2 clusters. $\Gamma$ consisted in changing angular distance to the diagonal by factor 0.05. $\Gamma^{++}$ with $\lambda=0.5$ was applied to a random subset of datapoints of size up to 1/3 of the dataset. 
  Default restart number was used for $k$-means. 
  No clustering error was observed for $\Gamma^{++}$ and up to 1\% of errors occurred for $\Gamma$ (mean errors 0.04\%,  std dev  0.007\%).  

\Bem{
Kleinberg errors
JKerrors
26 27 29 30 31 32 33 34 35 36 37 38 39 40 41 42 43 44 45 46 47 48 49 50 51 52 
 1  1  2  1  4  6 11  7 12 10 11  9 11 11 14 10 15  9  6 12  3  7  4  2  2  1 
53 54 55 56 57 58 60 61 62 81 
 3  4  1  1  2  3  1  1  1  1

mean errors 41.39 std dev= 7.49324 

Klvopotek errors
MKerrors
  0 
200 

berg errors
JKerrors
24 25 29 30 31 32 33 34 35 36 37 38 39 40 41 42 43 44 45 46 47 48 49 50 51 52 
 1  1  2  2  3  2  9 10 12 11 11 10 13 10 11 10 12 12 11  4  6  8  8  4  3  3 
53 54 55 56 58 68 70 
 1  5  1  1  1  1  1 

mean errors 41.31 std dev= 6.886291 

Klvopotek errors
MKerrors
  0 
200

Kleinberg errors
JKerrors
23 25 26 27 28 29 30 31 32 33 34 35 36 37 38 39 40 41 42 43 44 45 46 47 48 49 
 1  2  1  3  2  2  5  2  7  6  6  4 11 14  6  7 13  8 10  9 15 13  9  5 10  3 
50 51 52 53 54 55 56 59 61 62 66 
 3  5  3  5  2  3  1  1  1  1  1 

mean errors 41.33 ( 0.4133 

Klvopotek errors
MKerrors
  0 
200 

 Execution time
Time difference of 39.84035 secs

} 

\section{Final Remarks}\label{sec:final}
The result of this note not only provides with a new cluster-preserving data transformation for $k$-means-like algorithms, and hence providing with a new, more flexible test bed, but also suggests that the rigid Kleinberg's consistency axiom can be relaxed at least for this category of algorithms. 

Further research is needed in order to create
similar transformations suitable for other clustering algorithms and in this way to discover 
a more general axiomatic system for clustering algorithms. 

Though the goal of the proposed cluster preserving transformation was to provide with some new methods for generating test{}bed data for $k$-means like algorithms, its applicability in for example some brands of privacy preserving needs further investigation.

\vskip 0.2in
\bibliographystyle{plain}
\bibliography{bibliography_bib}

\end{document}